\pdfoutput=1
\documentclass[11pt]{article}

\usepackage{EACL2023}

\usepackage{listings}

\usepackage{times}
\usepackage{latexsym}

\usepackage[T1]{fontenc}
\usepackage[utf8]{inputenc}

\usepackage{microtype}

\usepackage{inconsolata}

\title{COCKATIEL: COntinuous Concept ranKed ATtribution with Interpretable ELements for explaining neural net classifiers on NLP tasks}

\author{Test \\ Address line \\  ... \\ Address line
      \And  ... \And
       Author n \\ Address line \\ ... \\ Address line}
\usepackage{amsthm}
\usepackage{amsmath}
\usepackage{amsfonts}
\usepackage{amssymb}
\usepackage{caption}
\usepackage{subcaption}
\usepackage{dsfont}
\usepackage{mathtools}
\usepackage{bbm}
\usepackage{bm}
\usepackage{multirow}
\usepackage{multicol}

\DeclareMathOperator*{\argmin}{arg\,min}

\theoremstyle{definition}
\newtheorem{definition}{Definition}[section]

\definecolor{blue}{RGB}{66, 133, 244}
\definecolor{red}{RGB}{219, 68, 55}
\definecolor{yellow}{RGB}{244, 188, 0}
\definecolor{green}{RGB}{15, 157, 88}
\definecolor{dark}{RGB}{31, 31, 31}
\definecolor{purple}{RGB}{113, 78, 163}
\definecolor{orange}{RGB}{187, 85, 39}
\definecolor{indigo}{RGB}{63, 81, 181}

\newcommand{\pred}{\bm{f}}

\newcommand{\vy}{\bm{Y}}
\newcommand{\vu}{\bm{U}}
\newcommand{\vw}{\bm{W}}

\newcommand{\va}{\bm{A}}
\newcommand{\vm}{\bm{M}}
\newcommand{\Var}{\mathbb{V}}

\newcommand\blfootnote[1]{%
  \begingroup
  \renewcommand\thefootnote{}\footnote{#1}%
  \addtocounter{footnote}{-1}%
  \endgroup
}

\author{Fanny Jourdan* \\
  IRIT, Université Paul-Sabatier  \\
  Toulouse, France \\
  \small \texttt{fanny.jourdan@irit.fr }
  \And
  Agustin Picard*\dag \\
  IRT Saint-Exupéry \\
  Toulouse, France \\ 
  \small \texttt{agustin-martin.picard@irt-saintexupery.com}
  \And
  Thomas Fel \\
  Brown University, USA \\
  SNCF, Toulouse, France \\\AND
  Laurent Risser \\
  IMT, Université Paul-Sabatier  \\
  Toulouse, France \\ \And
  Jean-Michel Loubes \\
  IMT, Université Paul-Sabatier  \\
  Toulouse, France \\ \And
  Nicholas Asher \\
  IRIT, Université Paul-Sabatier  \\
  Toulouse, France \\
  }

\begin{document} 
\maketitle
\begin{abstract}
Transformer architectures are complex and their use in NLP, while it has engendered many successes, makes their interpretability or explainability challenging.  Recent debates have shown that attention maps and attribution methods are unreliable \cite{pruthi2019learning, brunner2019identifiability}.  In this paper, we present some of their limitations and introduce COCKATIEL, which successfully addresses some of them. COCKATIEL is a novel, post-hoc, concept-based, model-agnostic XAI technique that generates meaningful explanations from the last layer of a neural net model trained on an NLP classification task by using Non-Negative Matrix Factorization (NMF) to discover the concepts the model leverages to make predictions and by exploiting a Sensitivity Analysis to estimate accurately the importance of each of these concepts for the model. It does so without compromising the accuracy of the underlying model or requiring a new one to be trained. 

We conduct experiments in single and multi-aspect sentiment analysis tasks and we show COCKATIEL's superior ability to discover concepts that align with humans' on Transformer models without any supervision, we objectively verify the faithfulness of its explanations through fidelity metrics, and we showcase its ability to provide meaningful explanations in two different datasets.
\end{abstract}

\begin{figure}
    \centering
\includegraphics[width=1\linewidth]{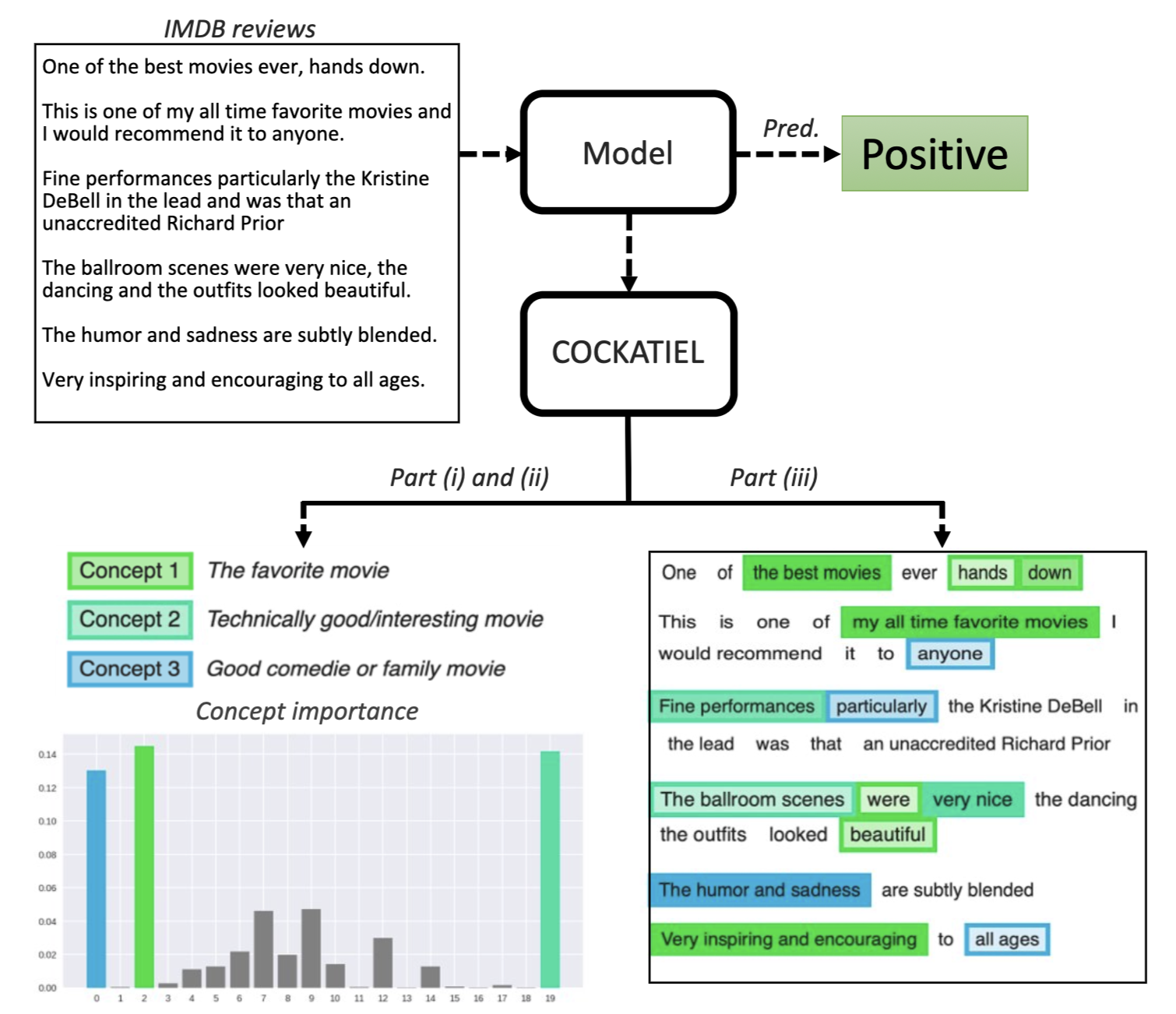}
\caption{\textbf{An illustration of COCKATIEL}. Given some sentences of IMDB reviews, COCKATIEL \textit{(i)} identifies concepts for prediction, \textit{(ii)} ranks them, and \textit{(iii)} gives the most important elements for each concept (to help us interpret the concept).}
\label{fig:visu_positivelabel}
\end{figure}

%%%%%%%%%%%%%%%%%%%%%%%%%
%%% TODO LIST
%%%%%%%%%%%%%%%%%%%%%%%%%
% - Figures:
%   - Résultats pour première page
%   - Résultats qualitatifs
\blfootnote{* Denotes equal contribution}
\blfootnote{\dag Work done as a Scalian employee, before April 2023 and joining IRT Saint-Exupéry.}

\section{Introduction}

NLP models have undeniably gotten increasingly more complex since the introduction of the transformer architecture~\cite{vaswani2017attention,devlin2018bert,liu2019roberta}. This trend, which is also occurring in the domain of Computer Vision, has brought about a need for understanding how these models make their predictions. The presence of bias in these models could indeed be prejudicial in applications where the user's lives are at stake~\cite{de2019bias}. 
Humans should be able to comprehend the reasons behind the model's decisions if these models are to gain general acceptance. Also, companies need to ensure that they are deploying algorithms which are free of harmful biases and that the explanations that they are obligated to issue are easily understandable by employees and end-users alike~\cite{kop2021eu}.  
%It is important to remark that this need for understanding automatic decisions start being enforced by Law, as for instance by the so-called \emph{AI act}\footnote{\url{https://eur-lex.europa.eu/legal-content/EN/TXT/?uri=celex\%3A52021PC0206}} of the European Union. As a consequence, companies need to ensure that they are deploying algorithms which are free of harmful biases and that the explanations that they're obligated to issue are easily understandable by employees and end-users alike~\cite{kop2021eu}.  

Intelligibility by humans has then become a key topic in explainable AI systems. As AI systems become more sophisticated and are deployed in increasingly complex environments, the ability to provide clear and concise explanations of their decisions becomes more pressing. %for building trust and ensuring the responsible use of these technologies.

Researchers have proposed multiple solutions to address this challenge. 
%of providing human interpretable explanations of complex AI models. 
The most straightforward approach analyzes how each part of the input influences the model's prediction.  There are  different ways of doing this, through perturbation~\cite{ribeiro2016model,zeiler2014visualizing} or by leveraging the gradients inside the neural network~\cite{sundararajan2017axiomatic}. However, these approaches suffer from being vulnerable to adversarial manipulation~\cite{wang2020gradient}, from only performing partial input recovery~\cite{adebayo2018sanity} and from a general lack of stability with respect to the input~\cite{ghorbani2019interpretation}. Another research path for transformer models harnesses the information in the attention maps of the transformers' layers to understand how the elements in the input relate to the output, implying that the attention mechanism is inherently interpretable. In spite of a number of supporters initially for this approach, there has been a recent wave of detractors of attention-based explanations~\cite{jain2019attention,pruthi2019learning,serrano2019attention}.

More in line with our proposal work, researchers in the field of rationalization have proposed specific architectures %based on bidirectional LSTMs 
to extract excerpts from whole inputs and predict a model's output based on these \emph{rationales}~\cite{lei2016rationalizing,jain2020learning,chang2020invariant,yu2019rethinking,bastings2019interpretable,paranjape2020information}. These rationales can be seen as explanations that are sufficiently high-level to be easily understood by humans. However, they require to train an entirely new model. Only one rationale can also be found per input text, when  there might intuitively be several predictions for a given prediction. Finally, these approaches use architectures that have mostly been left behind since the introduction of the transformer architecture, due to their inferior predictive capabilities.

In line with the project of generating explanations that are meaningful to humans, concept-based explainable AI (XAI) has lately advanced the sate of the art. The pioneer method TCAV~\cite{kim2018interpretability} goes beyond widespread attribution methods to create high-level explanations based on hand-picked concepts. More recently, \citet{fel2022craft} has extended this technique to discover automatically pertinent concepts inside the network's activation space and to find the parts of the input space that most align with each concept. Still, it has only been applied to convolutional architectures for image classification tasks.

In this paper, we present COCKATIEL, a novel technique for generating reliable and meaningful explanations for NLP neural architectures for classification problems.  It extends CRAFT~\cite{fel2022craft} and our contributions can be summarized as follows:

\begin{itemize}
    \item We introduce a post-hoc explainability technique that is applicable to any neural network architecture containing non-negative activation functions. The technique is capable of explaining predictions of individual instances as well as providing insights of the model's general behavior. % improve wording...
    %\item A methodology for soundly choosing the amount of concepts needed to faithfully explain models, effectively removing all the hyper-parameters from the method. %improve wording...
    \item We measure COCKATIEL's ability to discover concepts that align with those that Humans would employ in a sentiment analysis application. 
    %when it trained to predict the general sentiment of reviews. 
    Although we did not train the model on data annotated with these human concepts, COCKATIEL's explanations find them with high accuracy.
    %We show that although the neural networks weren't explicitly trained to learn these human concepts, they still needed to to be able to achieve high accuracy. {\color {magenta} que veux tu dire avec la derniere phrase?}
    %\item We demonstrate that, in addition to generating concepts that are meaningful to humans, these explanations are faithful to the models, {\color {purple} thus ensuring that we are not falling for cognitive biases on our predictions}.  {\color {magenta} peut être il faut une discussion de faithfulness.  Ça arrive comme ça.  En plus la derniere clause en pourpre n'aide pas à comprendre.}
    \item  We demonstrate that in addition to generating meaningful concepts for Humans, these explanations are faithful to the models: An explanation $X$ provided by method $C$ is faithful to a model $M$ just in case if $X$ is returned as a putative explanation of $M$'s behavior by $C$, the $X$ plays a causal role in $M$'s behavior. 
    \item We provide examples of explanations on fine-tuned RoBERTa models~\cite{liu2019roberta} and bidirectional LSTMs trained from scratch to show how the concept decomposition can be used to understand the inner workings of complex models.
    %\item \agus{commentaire sur les résultats qualitatifs ?}
\end{itemize}

\section{Related Work}

\subsection{Explaining through rationalization}

Finding rationales in text refers to the process of identifying expressions that provide the key reasons or justifications that are provided for a particular claim or decision about that text.   %This can be useful in a variety of contexts, such as natural language processing, information retrieval, and argumentation mining.  ce n'est pas necessaire
\citet{lei2016rationalizing} defined rationales as "a minimal set of text spans that are sufficient to support a given claim or decision". They should satisfy two desiderata: they should be interpretable, and they should reach nearly the same prediction as the original output. To do so, they use a generator network that finds interesting excerpts and an encoder network that generates predictions based on them.  However, their scheme requires the use of reinforcement learning~\cite{reinforce} for the optimization procedure. \citet{bastings2019interpretable} proposed to include a reparametrization trick to allow for better gradient estimations without the need for reinforcement learning techniques, and a sparsity constraint to encourage the retrieval of minimal excerpts.

%\citet{yu2019rethinking} looks at the problem of producing adequate rationales from a game-theoretic point of view.  Supposing we have a game with two players, where one player chooses information without a notion of the target {\color {magenta} c'est quoi le target?}, and the other having limited information from the input to explain the output {\color {magenta] c'est quoi le rôle du 2eme joueur?}, the resulting models can suffer from poor performance, and their control of the information at their disposal can be quite minimal. They propose to solve this by adding a third player that predicts the output based on a complement rationale (that is generated by the excerpt-generator, as well). \citet{paranjape2020information} on the other hand proposes to optimize an information bottleneck objective to better manage the trade-off between the amount of information contained in each rationale and the model's classification accuracy.{\color {magenta} cette description pour moi est assez confuse.  Peux tu définir proprement le jeu? C'est quoi les buts des deux joueurs et que fait le 3eme et comment interagit-t-il avec les 2 autres?}

\citet{yu2019rethinking} and \citet{paranjape2020information} studied the problem of producing adequate rationales from a game-theoretic point of view.  However, these models can be quite complex to train, as they either require a reparametrization trick or a reinforcement learning procedure. \citet{jain2020learning} proposed to solve this problem by introducing a support model capable of producing continuous importance scores for instances of the input text, that the rationale extractor can  use to decide whether an excerpt will make  a good rationale or not.

All these rationales will serve as an explanation for single instances, but won't explain how models predict whole classes. \citet{chang2019game} introduced a rationalization technique that allows for the retrieval of rationales for factual and counterfactual scenarios using three players.

However, all these techniques are not model agnostic and require specific architectures, in particular rather simple architectures %and 
or LSTMs, and training procedures.  But these architectures have been shown to not produce optimal results. 

%and produce less accurate predictions.
%{\color {magenta} t'as une citation pour cette critique ou as tu des scores plus bas pour ça?  Il faut préciser}

\subsection{Concept-based explanations}

Concept-based explainability is a growing area of research in AI, focused on generating human-understandable explanations for the decisions made by machine learning models. One popular approach for generating concepts is TCAV~\cite{kim2018interpretability}.
It uses gradient-based techniques to identify the important features of a model. However, TCAV relies on Human inputs, as it requires the user to manually specify the concepts to be tested. This can be time-consuming and may not always produce the most comprehensive explanations \cite{ghorbani2019towards}.  

Another approach, ACE~\cite{ghorbani2019towards}, aims to automate the concept extraction process. ACE uses a clustering algorithm to identify interpretable concepts in the model's activations, without the need for Human input. While this approach has the potential to greatly reduce the time and effort required for concepts extraction, the authors criticize their own reliance on pre-defined clustering algorithms, which may not always produce the most relevant or useful concepts.
%While this approach has the potential to greatly reduce the time and effort required for concept extraction, it has been criticized for its reliance on pre-defined clustering algorithms, which may not always produce the most relevant or useful concepts.{\color {magenta} citation pour cette critique?}

An alternative uses matrix factorization techniques, such as non-negative matrix factorization (NMF)~\cite{lee1999learning}, to identify interpretable factors in the data~\cite{zhang2021invertible,fel2022craft}. As presented in Section~\ref{sec:cockatiel}, or strategy is inspired by~\cite{fel2022craft} and is therefore a concept-based explanations XAI method.
In \cite{fel2022craft}, the authors developed a framework for generating global and local explanations. They successfully tested the meaningfulness and the capacity of these explanations to help Humans to understand the model's behavior through psychological experiments. 
%{\color {magenta} peut etre il faut dire un peu plus la comme Cockatiel est proche de craft---mentionner les points forts}  
However, this approach has only been applied to convolutional neural networks for image classification tasks so far.

For NLP applications, \citet{bouchacourt2019educe} proposed a self-interpretable neural architecture capable of simultaneously generating a prediction on classification tasks and its concept-based explanation. These concepts are learned without supervision from excerpts using a bidirectional LSTM during the training phase of the model, and the predictions are only based on the presence or absence of the individual concepts in the input sentences.  Despite of its capacity to generate interesting concepts, its low prediction accuracy for the classification task is a serious limitation (see Table 1). %LR: A appuyer d'une reference et/ou a serieusement temperer. On ne peut pas attaquer un papier sans argument solide... et il est en general deconseille d'attaquer un papier de maniere si frontale, car un de ses auteurs pourra etre ton reviewer. 
%Je peux citer directement la partie 4 du papier où on les bats énormément sur le tableau des scores?
Going further, \citet{antognini-faltings-2021-rationalization} introduced ConRAT, a technique that includes orthogonality, cosine similarity and knowledge distillation constraints, as well as a concept pruning procedure to improve on both the quality of the extracted concepts and the model's accuracy.% Quelle est la limitation sur ce papier?

\begin{figure*}[t]
    \centering
    \includegraphics[width=\textwidth]{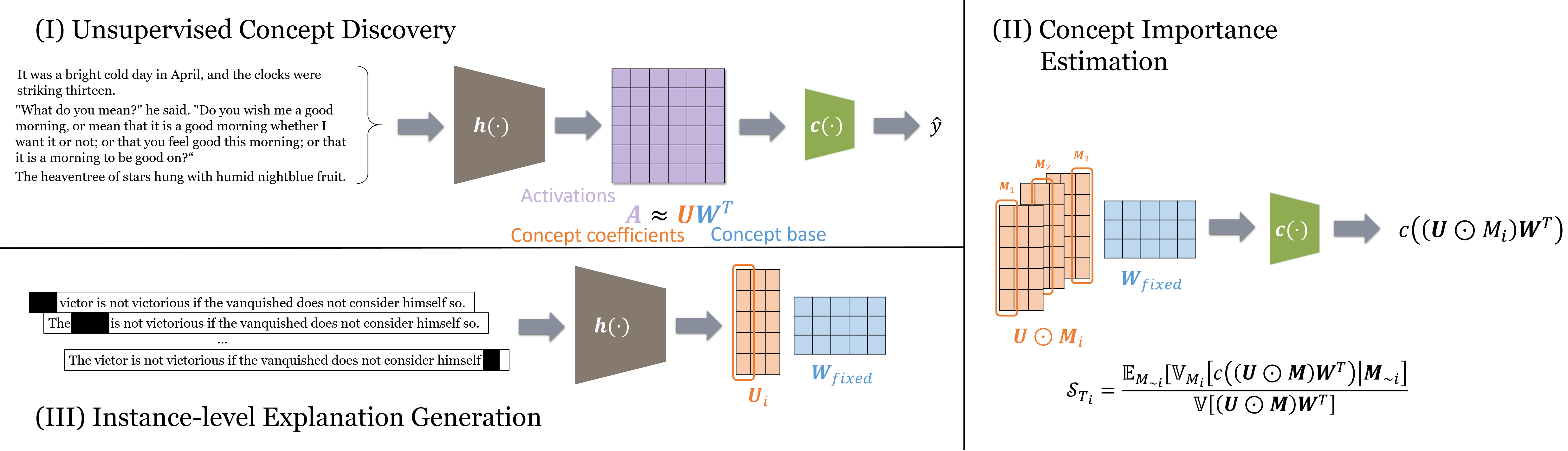}
    \caption{\textbf{Overview of our method:} COCKATIEL can be divided into three phases. \textit{(i)} The first step is assembling the concepts base. We propose to do this by constituting a database of whole or excerpts of input texts, projecting each one of these elements into the embedding of the model of our choice $h (\bm{x})$ and using the NMF algorithm to decompose the resulting non-negative matrix into two low-rank, non-negative matrices: $\vu$ and $\vw$. \textit{(ii)} Once $\vu$ and $\vw$ have been computed, we can compute the Total Sobol indices for the concept base's columns by masking the coefficients and by looking at their effect on the classifier's output: $c((\vu \odot \vm)\vw^T)$. \textit{(iii)} Finally, we propose to retrieve the influence of each word of the instance under study in each concept through Occlusion, that is, by applying masks to each word (or clause) in the input and quantifying the changes in each of the concept coefficients.}
    \label{fig:diagram}
\end{figure*}

\section{COCKATIEL}\label{sec:cockatiel}

In this section, we describe COCKATIEL, our concept-based XAI technique for NLP models to generate human-understandable explanations.  It has three main components: \textit{(i)} it uses Non-Negative Matrix Factorization (NMF) to discover the concepts that the neural network under study leverages to make predictions; \textit{(ii)} it exploiting Sensitivity Analysis to estimate accurately the importance of each of these concepts for the model; and \textit{(iii)} it uses a black-box explainability technique to generate instance-wise explanations at a per-word and per-clause level.  Fig.~2 presents a schematic outline of COCKATIEL.

\paragraph{Notation}

In a supervised learning framework, we assume that a neural network model $\pred \colon \mathcal{X}^n \rightarrow \mathcal{Y}^n$ has already been trained for some classification task. We denote by $(\bm{x}_1, ..., \bm{x}_n) \in \mathcal{X}^n$ a set of $n$ input texts and $(y_1, ..., y_n) \in \mathcal{Y}^n$ their associated labels.  %$\pred$ predicts the output $\pred(\bm{x})$ at some test input $\bm{x}$.  
We consider $\pred$ to be a composition of $h$, the last embedding of $\bm{x}$ (\textit{i.e.} the last layer of the feature extractor model), and $c$, the classification function, 
$\pred(\bm{x}) = c \circ h(\bm{x})$ with $ h(\bm{x}) \subseteq \mathbb{R}^p$ .

 COCKATIEL will factorize $h$ through NMF, so we require $h$ to be non-negative -- i.e. $h(\bm{x}) \geq 0 \; \forall \bm{x} \in \mathcal{X}$. This constraint is typically verified when the last layer has an activation function such that $\bm{\sigma}(\bm{x}) \geq 0$, which is the case in (but it's not limited to) layers or blocks using \textit{ReLU}.

%We consider this neural network to be a composition of $k$ layers, where $\pred(\bm{x}) = \vh_{k} \circ \vh_{k-1} \circ ... \circ \vh_{1}(\bm{x})$ with $\vh_l(\bm{x}) \subseteq \mathbb{R}^p$ denotes intermediate activations for the layer $l$, and $\vh_{l}(\bm{x})_i$, an activation for that same layer. As we will be factorizing these activations through the NMF algorithm, we require them to be non-negative -- i.e. $\vh_l(\bm{x})_i \geq 0 : \forall i \in \{1, ..., p\}$. This constraint is typically verified by choosing layers whose activation function $\bm{\sigma}_l(\bm{x}) \geq 0$, which is the case for (but it's not limited to) layers (or blocks) using \textit{ReLU}. Finally, we define the top-side of the network going from the desired activations to the model's output as $\vh_{l \to k}$ -- e.g. $\pred(\bm{x}) = \vh_{j \to k} \circ \vh_{j}(\bm{x}), ~ 1 < j < k$.

\subsection{Unsupervised concept discovery - "Concept part"} % NMF + choice of \tau

COCKATIEL discovers concepts without supervision by factorizing the neural network's intermediary activations by using a NMF algorithm. %In~\cite{fel2022craft}, the authors explain that this choice was made empirically, as the best results were given by this dimensionality reduction technique. However, in the case of NLP, this choice is perfectly justified by the literature. \agus{Rajouter les justifications et les bonnes réfs !}
Because we are factorizing $h$, we can generate explanations on embeddings without needing to deal with the complexities of attention layers~\cite{pruthi2019learning}; nor do we have to deal with the non-identifiability of transformer models~\cite{brunner2019identifiability}. Thus, the concept extraction phase of our method does not depend on the specificities of attention. We will address this later on in Section~\ref{sec:local-explanations} to be able to generate our instance-level explanations.  

% In the methods for explaining transforming models, a majority are constructed in such a way that they use the attention of these models.
% A first category of method directly uses attention as an explanation. These types of methods are no longer used because \citet{pruthi2019learning} has shown that they are not reliable.  For this, they create a method for generating attention-based explanations that are designed to mislead humans, and show that these explanations can be effective at deceiving humans.
% The second category of method are those that directly explain the output with the input of the model (so the words / sentences directly) like LIME \cite{LIME}. These methods are not optimal because \citet{brunner2019identifiability} shows that the transformer-based language models are not identifiable. That means that there can be multiple sets of parameter values that yield the same output for a given input.

% The NMF allows to have an explanation in relation to the output embedding of the last layer of the encoding and not the input of the model. It is therefore independent of attention.

\paragraph{NMF algorithm:}
We choose an excerpt-extraction function $\bm{\tau}_1$ to generate a database of excerpts coming from texts that the model places in the desired class $d_c$ -- i.e. $\bm{X}_i = \bm{\tau}_1 (x_i)$ such that $\pred(x_i) = d_c$. 
%Then, we pick a layer $l$ with non-negative activation function and extract the activations $\bm{A} = \vh_{l \to k} (\bm{X}_i)$ for each of the excepts $\bm{X}_i$ in the database, and to solve the constrained optimization problem engendered by the NMF algorithm:
Then, we place ourselves at the model's last layer %(which we require to have a non-negative activation function) 
and we extract the activations $\bm{A} = h (\bm{X}_i)$ for each of the excerpts $\bm{X}_i$ in the database. With this information, we solve the constrained optimization problem engendered by the NMF algorithm:
\begin{equation}
    \label{eq:nmf}
    (\vu, \vw) = \argmin_{\vu \geq 0, \vw \geq 0} ~ \frac{1}{2}\|\va - \vu\vw^T\|^2_{F}, %\quad\textcolor{red}{\text{non convex, biconvex, NP hard}}
\end{equation}  
where $||\cdot||_F$ is the Frobenius norm.

This allows us to decompose the high-rank matrix containing all activations $\va \in \mathbb{R}^{n \times p}$ into two low-rank matrices $\vu \in \mathbb{R}^{n \times r}$ and $\vw \in \mathbb{R}^{p \times r}$. Intuitively, this corresponds to $\vw$ being a matrix whose columns represent the concepts that we will use to generate explanations, and $\vu$ is a matrix containing the coefficients quantifying the presence of each concept. These matrices are built so as to minimize the reconstruction error $\frac{1}{2}\|\va - \vu\vw\|^2_F$, enforcing the relevance of the concepts, and with a non-negative constraint for each matrix, thus encouraging sparsity in their elements.

It is important to note that these coefficients $u_{ij} \in \mathbb{R}_+$, so the presence of a concept can be determined by where its value stands in the concept's coefficients distribution. In practice, we have found that fixing a threshold at the quantile representing the 10\% highest values leads to accurate and easy to interpret explanations.

\paragraph{Choice of $\bm{\tau}_1$:} 
As we want the concepts to be descriptive enough to convey an abstraction but short enough to only contain one, we work with excerpts chosen by an excerpt-extraction function $\bm{\tau}_1$. The choice of $\bm{\tau}_1$, which should depend on the dataset and the text's format,  heavily impacts the type of explanations that we are able to generate.

We have identified 3 possible $\bm{\tau}_1$ functions: \textit{(i)} take all the full text ; \textit{(ii)} split the text into sentences (of at least 6 words) ; \textit{(iii)} split the text into clauses. Linguistically, it doesn't make sense to take smaller tokens like one or two words since their meaning is typically too unfocused to provide a real explanation.

We therefore chose $\bm{\tau}_1$ to respond specifically to each use-case. If we want to capture the mood of whole inputs, we can designate the inputs as the excerpts, and then interpret them by leveraging the local part of our method. If we instead wish to extract more simple but structured concepts, we can choose $\bm{\tau}_1$ to pick sentences of at least 6 words and ending in a full-stop. The first condition is necessary in the case of the \textit{beer review} dataset, which is composed of short sentences containing very simple descriptions.  For this dataset, using only very short excerpts would fail to convey the complexity of the ideas conveyed by the concepts. In this paper, we present results using these two excerpt-extraction functions. %, but other smarter choices could be envisaged.

\begin{figure}[t]
    \centering
\includegraphics[width=1\linewidth]
{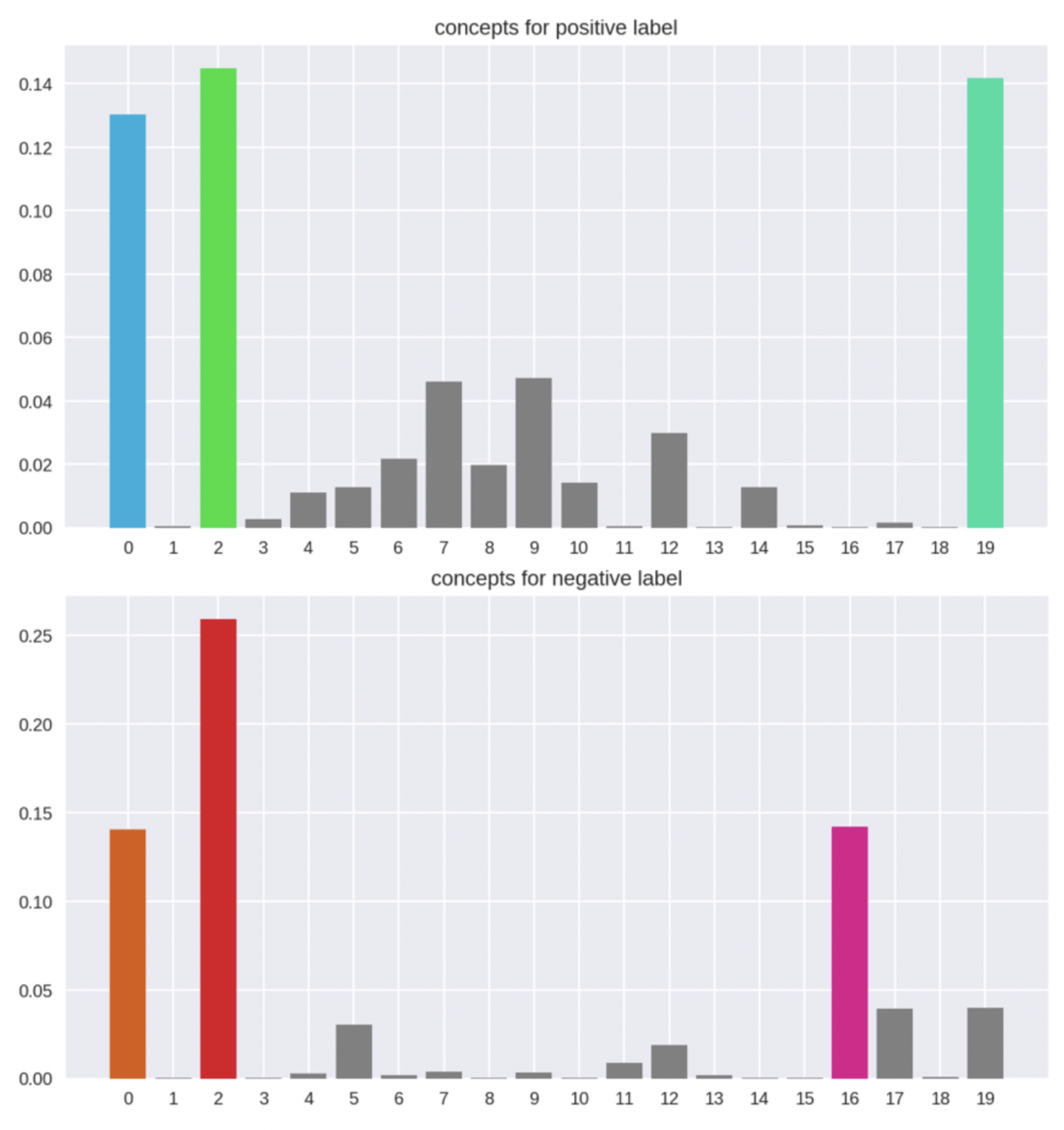}
    \caption{\textbf{Concept importance: } The global influence of the NMF concepts on the predictions on RoBERTa model is  measured using Sobol indices. There are different concepts for each class (positive and negative label).}
    \label{fig:importanceconcept_roberta}
\end{figure}

\subsection{Concept importance estimation - "Ranking part"} % Sobol

%Following the inspiration in \cite{fel2022craft}, 
 A common issue when utilizing concept extraction methods is the discrepancy between concepts deemed relevant by humans and those utilized by the model for classification. To mitigate the potential for confirmation bias during the concept analysis phase, we estimate the overall importance of the extracted concepts.

To determine which concept has the most significant impact on the model output, we use a counterfactual reasoning \cite{peters2017elements, pearl2016causal}, and then use sensitivity analysis \cite{cukier1973study, iooss2015review}. A classic strategy in this area is the use of total Sobol indices \cite{sobol1993sensitivity}.
This method captures the importance of a concept, along with its interactions with other concepts, on the model output by calculating the expected variance that would remain if all the indices of the masks except $\vm_i$ were fixed. 

\begin{definition}[\textbf{Total Sobol indices}]
\textit{The total Sobol index $\mathcal{S}{T_i}$, which measures the contribution of a concept $\vu_i$ as well as its interactions of any order with any other concepts to the model output variance, is given by:}
\begin{align}
        \label{eq:total_sobol}
           \mathcal{S}_{T_i} 
           & = \frac{ \mathbb{E}_{\bm{M}_{\sim i}}( \Var_{M_i} ( \vy | \bm{M}_{\sim i} )) }{ \Var(\vy) } \\
           & = \frac{ \mathbb{E}_{\bm{M}_{\sim i}}( \Var_{M_i} ( c((\vu \odot \vm)\vw^T) | \bm{M}_{\sim i} )) }{ \Var( c((\vu \odot \vm)\vw^T)) }.
    \end{align}
\end{definition}

To estimate the importance of a concept $\vu_i$, we measure the fluctuations of the model output $c(\vu \vw^T)$ in response to perturbations of the concept coefficient $\vu_i$. Specifically, we use a sequence of random variables $\vm$ to introduce concept fluctuations and reconstruct a perturbed activation $\tilde{\va} = (\vu \odot \vm)\vw^T$. We then propagate this perturbed activation to the model output $\vy = c(\tilde{\va})$. An important concept will have a large variance in the model output, while an unused concept will barely change it.

The method for calculating (2) and (3) exploits the Sobol-Hoeffding decomposition  and is in the supplementary materials (appendix~\ref{apdx:sobol}).

There are already a plethora of different techniques that allow us to compute this index efficiently~\cite{saltelli2010variance, marrel2009calculations, janon2014asymptotic, owen2013better, tarantola2006random}. But concretely, we estimate the total Sobol indices using the Jansen estimator \cite{janon2014asymptotic}, a widely recognized efficient method \cite{puy2022comprehensive}. 
The Jansen estimator is commonly utilized in conjunction with a Monte Carlo sampling strategy, but we improve over Monte Carlo by using a Quasi-Monte Carlo sampling strategy. This technique generates sample sequences with low discrepancy, resulting in a more rapid and stable convergence rate \cite{gerber2015integration}.

\subsection{Instance-level explanation generation - "Interpretable elements part"}\label{sec:local-explanations} % Local, implicit diff

In this part, we interpret the concepts found previously. To do this, we find which words and clauses are associated with each concept.

We adapt Occlusion~\cite{zeiler2014visualizing}: a black-box attribution method that works by masking each word looking at the impact on the model output. In this case, to get an idea of the importance of each word for a given concept, we mask words in a sentence and measure the effect of the new sentence (without the words) on the concept. This operation can be performed at word or clause level -- i.e. mask words or whole clauses -- to obtain explanations that are more or less fine-grained depending on the application.

\paragraph{Motivations:}

This choice has been shown to perform particularly well on NLP models~\cite{fel2021sobol} and doesn't suffer from the inefficiency of having to sample a considerable amount of masks for each explanation. 
Indeed, in ~\cite{fel2021sobol}, they compared Occlusion to other explainability techniques that are commonly used in NLP, and they showed that it is more faithful to the model than Saliency~\cite{saliency}, Grad-Input, SmoothGrad~\cite{smilkov2017smoothgrad}, Integrated Gradients~\cite{IntegratedGradients}, and their own Sobol method on both LSTM and BERT models.

In addition, in the case of transformer models, using a black-box method such as Occlusion avoids manipulating the attention layers between the input and the activation matrix $A$, where our concepts are located. In doing so, we avoid the non-identifiability problem of transformer models \cite{pruthi2019learning}.

\paragraph{Application:}
Empirically, we perform the following operations:

For a sentence $X_i$, $A_i = h(X_i)$. We have a fixed $W$ calculate with the NMF and $W_k$, the $k$ concept of $W$. As before, we get the importance of the sentence $X_i$ for the concept $k$:
\begin{equation*}
    U_i^k = \argmin_{U \geq 0} ~ \frac{1}{2}\| A_i - U W_k^T\|^2_{F}.
\end{equation*}  
Then, we remove the element $j$ from the sentence $i$: $\Tilde{X}_{i-j}$ (i.e. we replace the (tokenized) feature by a zero). So we have $\Tilde{A}_{i-j} = h(\Tilde{X}_{i-j})$, and:
\begin{equation*}
    \Tilde{U}_{i-j}^k = \argmin_{U \geq 0} ~ \frac{1}{2}\| \Tilde{A}_{i-j} - U W_k^T\|^2_{F}, 
\end{equation*}  

So, $\phi(k,i,j)$ quantifies the influence of the element $j$ in the sentence $i$ for the concept $k$:
\begin{equation*}
   \phi(k,i,j) =  U_i^k - \Tilde{U}_{i-j}^k, 
\end{equation*} 

For the visualisations (see e.g. Fig.~\ref{fig:examples}), we color the element with the color of the concept for which it is most important. In addition, the darker the color, the more important the element is for the concept.

% \fanny{
% \paragraph{Choice of $\bm{\tau}_2$:} 
% As for the NMF (part 1), the crops must be descriptive enough to understand the concept but short enough to contain only it that is readable. Then, we work with excerpts chosen by an excerpt-extraction function $\bm{\tau}_2$.
% The choice of $\bm{\tau}_2$, depends on the dataset, the text's format, and $\bm{\tau}_1$ (excerpt chosen by $\bm{\tau}_2$ must be contained in excerpt chosen by $\bm{\tau}_1$).
% It is important to look at different $\bm{\tau}_2$ functions to choose the most understandable one to optimize our explanation. To understand this trade-off, the figure \ref{fig:visu_tau} in appendix, contains visualization examples with different $\bm{\tau}_2$.
% }

\paragraph{Choice of $\bm{\tau}_2$:}
Just like in the case of the NMF, the choice of the form of the elements of the input to occlude will have an impact on the understandability of the explanations. This can be generalized via another excerpt extraction function $\bm{\tau}_2$, whose optimal shape will depend on the dataset, the text's format and the learned concepts (i.e. Occlusion shouldn't be applied at a per-clause level if the concepts were learned using a $\bm{\tau}_1$ providing single words, so this first exceprt extraction function must be taken into consideration). There is a certain trade-off between the granularity and the interpretability of the explanations, as illustrated in Figure~\ref{fig:visu_tau} in the appendix which contains some examples with different choices of $\bm{\tau}_2$. In general, we advise to try different combinations of $\bm{\tau}_i$ to find the desired level of granularity in the explanations for each use-case.

\begin{table*}[t]
%\label{tab:annotation_alignment}
\centering
\begin{tabular}{p{0.35cm}*{40}
{@{\hskip.9mm}c@{\hskip.9mm}}}
\label{tab:annotation}
& &  & \multicolumn{3}{c}{\textit{Average}} & \multicolumn{3}{c}{\textit{Appearance}} & \multicolumn{3}{c}{\textit{Aroma}} & \multicolumn{3}{c}{\textit{Palate}} & \multicolumn{3}{c}{\textit{Taste}} \\
\hline
& \textbf{Model} & \textbf{Acc.} & \textbf{Prec.} & \textbf{Rec.} & \textbf{F1} & \textbf{P} & \textbf{R} &  \textbf{F} & \textbf{P} & \textbf{R} & \textbf{F} & \textbf{P} & \textbf{R} & \textbf{F} & \textbf{P} & \textbf{R} & \textbf{F} \\
\hline
\multirow{6}*{\rotatebox{90}{$l = 20$}}  & RNP & 81.1 & 24.7 & 21.3 & 24.9 & 28.6 & 23.2 & 26.5 & 22.1 & 21.0 & 21.5 & 17.7 & 24.1 & 20.4 & 28.1 & 16.7 & 20.9 \\
& RNP-3P & 80.5 & 26 & 21.8 & 23.3 & 30.4 & 25.6 & 27.8 & 19.3 & 20.4 & 19.8 & 10.3 & 12.0 & 11.1 & 43.9 & 28.4 & 34.5 \\
& Intro-3P & 85.6 & 21 & 18.0 & 19.1 & 28.7 & 24.8 & 26.6 & 14.3 & 14.4 & 14.3 & 16.6 & 19.3 & 17.9 & 24.2 & 13.6 & 17.4 \\
& InvRAT & 82.9 & 37.5 & 31.6 & 33.8 & 54.5 & 45.5 & 49.6 & 26.1 & 27.6 & 26.9 & 22.6 & 25.9 & 24.1 & 46.6 & 27.4 & 34.5 \\
& ConRAT & \underline{91.4} & \textbf{43.8} & \underline{39.7} & \underline{40.9} & \underline{57.8} & \underline{53.0} & \underline{55.3} & \underline{31.9} & \underline{35.5} & \underline{33.6} & \textbf{29.0} & \underline{36.3} & \underline{32.3} & \textbf{56.5} & \underline{33.9} & \underline{42.4} \\
& \textbf{Ours} & \textbf{95.2} & \underline{40.6} & \textbf{58.4} & \textbf{47} & \textbf{67.5} & \textbf{71.4} & \textbf{69.4} & \textbf{34.1} & \textbf{42.3} & \textbf{37.7} & \underline{24.8} & \textbf{46.7} & \textbf{32.4} & 36.1 & \textbf{73.3} & \textbf{48.4} \\
\hline
\multirow{6}*{\rotatebox{90}{$l = 10$ }} & RNP & 84.4 & 32.7 & 14.5 & 19.5 & 40.1 & 12.0 & 18.5 & \underline{33.3} & \underline{18.7} & \underline{24.0} & \underline{25.1} & \underline{17.4} & \underline{20.6} & 32.3 & 9.8 & 15.07 \\
& RNP-3P & 83.1 & 28.4 & 13.2 & 17.8 & 41.8 & 19.2 & 26.3 & 22.2 & 12.4 & 15.9 & 16.5 & 10.4 & 12.7 & 33.2 & 10.6 & 16.1 \\
& Intro-3P & 80.9 & 24 & 12.2 & 16.1 & 51.0 & 26.0 & 34.4 & 18.8 & 9.7 & 12.8 & 16.5 & 10.6 & 12.9 & 9.7 & 2.6 & 4.1 \\
& InvRAT & 81.9 & 36.6 & 15.7 & 21.8 & \underline{59.4} & 26.1 & \underline{36.3} & 31.3 & 15.5 & 20.8 & 16.4 & 9.6 & 12.1 & 39.1 & 11.6 & 17.9 \\
& ConRAT & \underline{91.3} & \underline{38.2} & \underline{17.6} & \underline{23.8} & 51.7 & \underline{26.2} & 34.8 & \textbf{32.6} & 17.4 & 22.7 & 23.0 & 13.8 & 17.3 & \textbf{45.3} & \underline{13.1} & \underline{20.3} \\
& \textbf{Ours} & \textbf{95.2} & \textbf{39.5} & \textbf{58.4} &\textbf{ 45.5} & \textbf{63.3} & \textbf{56.4}  &  \textbf{59.7} & 27.3 & \textbf{67.4} & \textbf{38.9} & \textbf{26} & \textbf{43.5} & \textbf{32.5} &  \underline{41.4} & \textbf{66.1} & \textbf{50.9}\\
\end{tabular}
\caption{Objective performance of rationales for the multi-aspect beer reviews. All baselines are trained separately on each aspect rating, except for ConRAT~\cite{antognini-faltings-2021-rationalization}, which is trained on the \textit{Overall} label just like our method. Bold and underline denote the best and second-best results, respectively.}
\end{table*}

\section{Experimental evaluation}
For all of our results, we fine-tuned RoBERTa~\cite{liu2019roberta} based models on each dataset. We ensured the non-negativity of at least one layer of the model by adding a ReLU activation after the first layer of the 1-hidden-layer, dense MLP of the classification head. For the qualitative analysis, we  also tested COCKATIEL's performance on bidirectional LSTM models trained from scratch. More details about the implementations are left in appendix~\ref{apdx:finetuning}.

We will first analyze the meaningfulness of the discovered concepts by measuring their alignment with human annotations on the different aspects of a multi or single-aspect sentiment analysis task. Then, we will ensure that our explanations are faithful to the model through an adaptation of the insertion and deletion metrics to concept-based XAI. Finally, we will showcase some examples of explanations and of applications for our method.

\subsection{Alignment with human concepts}

\begin{figure}[h]
    \centering
\includegraphics[width=0.5\textwidth]{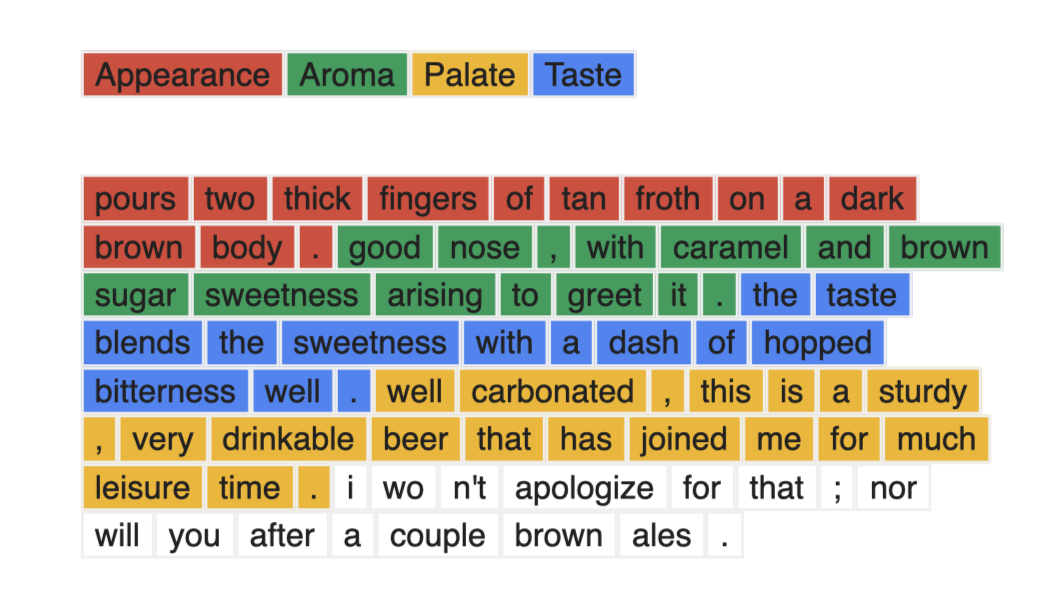}    
    \caption{Concepts generated with $l=20$ for a beer review. The colors depict the aspects for each annotate concept. COCKATIEL is trained only on the label and we use the NMF part of the method to find annotate concepts. \textit{For other examples of review, see appendix \ref{apdx:examples_sup}}}
    \label{fig:beer_visu}
\end{figure}

Following the human-alignment evaluation in~\cite{antognini-faltings-2021-rationalization}, we perform beer task:

\paragraph{Beer Task}
We will measure the extent to which our concepts overlap the human annotations for the 4 different aspects of the multi-aspect \textit{beer reviews} dataset~\cite{beer-reviews}. This dataset contains reviews for beers with commentary and marks (from 0 to 5) on 5 different aspects: Appearance, Aroma, Palate, Taste and Overall. The model will be trained to predict whether the overall score is greater than 3 -- i.e. a positive review on the beer -- and will not have access to the labels for the other aspects. Additionally, it includes 994 reviews with annotations indicating the position of these aspects in the text. The objective of this evaluation is to look for concepts that align with these annotations and measure their capacity to predict the location of each different aspect. In particular, we searched across the whole annotated dataset for the concepts whose F1 score for the prediction of each aspect was maximal. It is important to note that this does not take into account to which extent they are important for the model to predict, but this only serves as an automatized test for determining whether the explainability technique is capable of generating understandable concepts.

We calculate the precision, recall and F1 scores for each aspect, and we do so with $l = 10$ and $l=20$ concepts. We remind the reader that, unlike the baselines, our method is a post-hoc technique, and thus, the model does not need to be re-trained, and that changing the number of concepts takes only a few minutes of compute on GPU.

%\paragraph{Amazon Electronics task}
%The second task is in the single-aspect \textit{Amazon Electronics} dataset \cite{ni2019justifying}. We randomly picked 24,000 balanced samples with ratings of four and above or two and below and we binarized the ratings $\leq 2$ as negative and $\geq 4$ as positive. For create human annotations, we use \cite{chang2019game} procedure: we considered the first 50 tokens after the words “pros:” and “cons:” as the rationale annotations for the positive and negative labels, respectively. We create 471 annotations... We want finding rationales that align with human annotations. \fanny{a finir}
%\paragraph{}

% the table* breaks the reference, but is the only environment capable of making tables that look right in double column
In Table 1, %and \ref{tab:annotation_amazon}
we present a comparison of our results to those obtained with some rationalization techniques: RNP~\cite{lei2016rationalizing}, RNP-3P~\cite{yu2019rethinking}, InvRAT~\cite{chang2020invariant} %CAR~\cite{chang2019game} 
and ConRAT~\cite{antognini-faltings-2021-rationalization} for the task on \textit{Beer}. % and on \textit{Amazon Electronics}. 
We demonstrate that not only our model achieves the highest accuracy, but also that it outperforms all the other methods in its ability to accurately recognize the human annotations, be it by its precision, recall or F1 score.

% \begin{table}
% \centering
% \begin{tabular}{lcccc}
% \hline
% \textbf{Model} & \textbf{Acc.} & \textbf{P} & \textbf{R} & \textbf{F}\\
% \hline
% RNP & 75.5 & 32.6 & 18.8 & 23.8 \\ 
% RNP-3P & 70.0 & 49.4 & 28.4 & 36.0 \\ 
% Intro-3P & 75.2 & 22.1 & 12.8 & 16.2 \\ 
% InvRAT & 71.5  & 44.3 & 25.5 & 32.4 \\ 
% ConRAT-1 & 75.5 & \textbf{56.4 }& 32.5 & 41.3 \\ 
% \textbf{Ours} & \textbf{96.4} & 28.7 & \textbf{99.5} &\textbf{ 44.6} \\ 
% \hline
% \end{tabular}
% \caption{Accuracy and objective performance of rationales in automatic evaluation for the \textit{Amazon Electronics} dataset. Bold denote the best result.}
% \label{tab:annotation_amazon}
% \end{table}

\subsection{Evaluation of Explanation Faithfulness}

We have demonstrated that we can generate concepts that greatly align with humans', but to legitimately serve as an explainability technique, we must also guarantee its faithfulness. 
This element is key, as the concepts leveraged by the model may not perfectly align with humans in every task, but we still want the explanation to reflect what the model is doing.
% \fanny{We want explain the model and the concepts used by model are not necessary the same used by humans for our task.}
An XAI method is said to be faithful if its explanations faithfully convey the information that the model is using to generate its predictions. %{\color {magenta} this definition of faithfulness needs to go earlier in the paragraph right before related work , p. 2 second bullet} 
In~\cite{ghorbani2019towards,zhang2021invertible}, they proposed to use an adaptation of the deletion and insertion explainability metrics to concept-based methods. In essence, they proposed to gradually mask/add the concepts (following their importance) and seeing the impact on the logits. If the concepts are indeed important for the model to predict, they should drastically decrease/increase as vital information for the prediction is progressively being erased/added.

To evaluate the explanation Faithfulness and present qualitative results, we used the IMDB dataset \cite{maas-EtAl:2011:ACL-HLT2011}. The IMDB dataset is a collection of 50K movie reviews from the Internet Movie Database (IMDB) website. For each review, IMDB specifies whether it is positive or negative (the label). The dataset is balanced, with 25K positive and 25K negative reviews.
We used a RoBERTa model to predict the label from the reviews.

\begin{figure}[h]
    \centering
\includegraphics[width=0.4\textwidth]{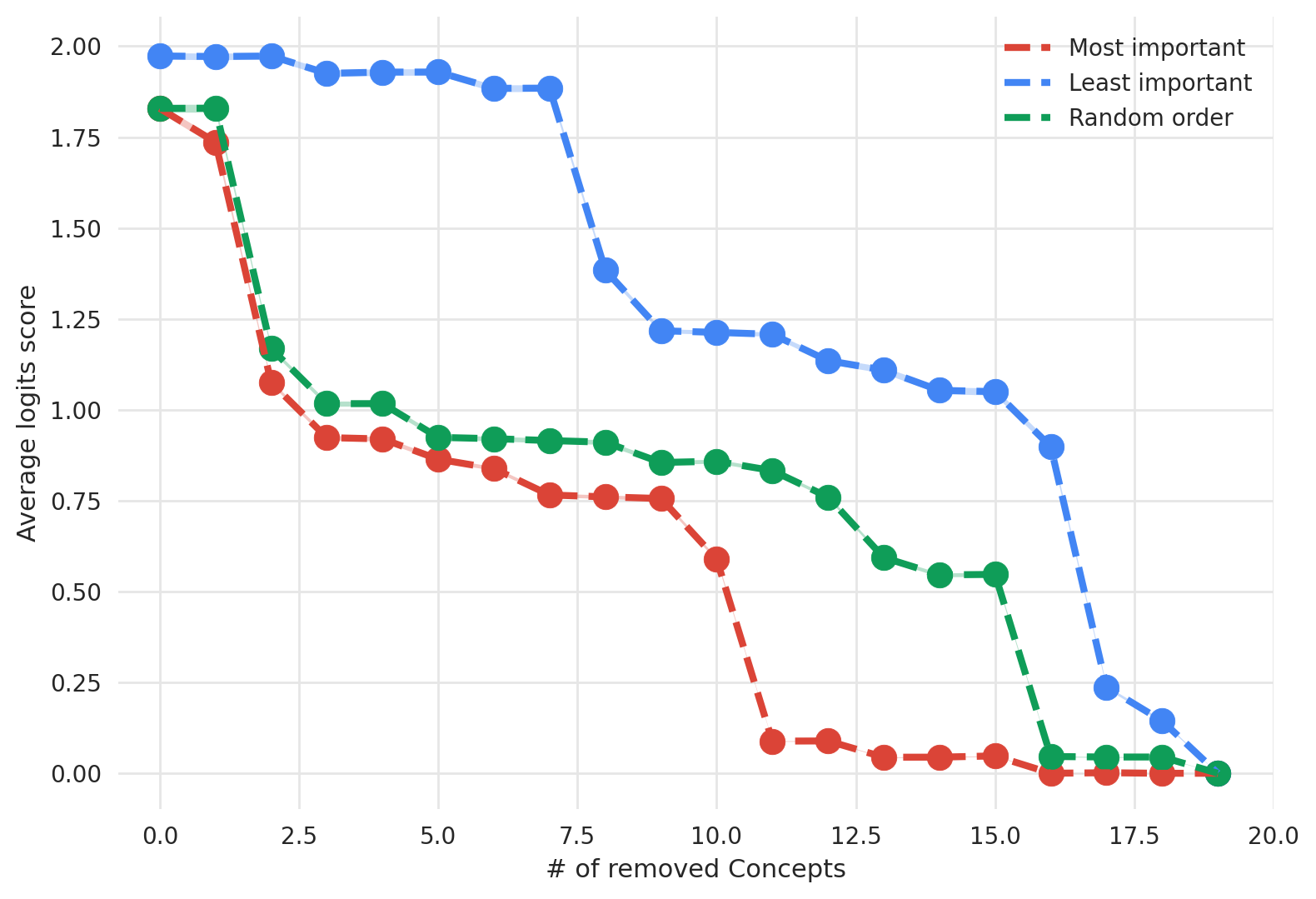}
\includegraphics[width=0.4\textwidth]{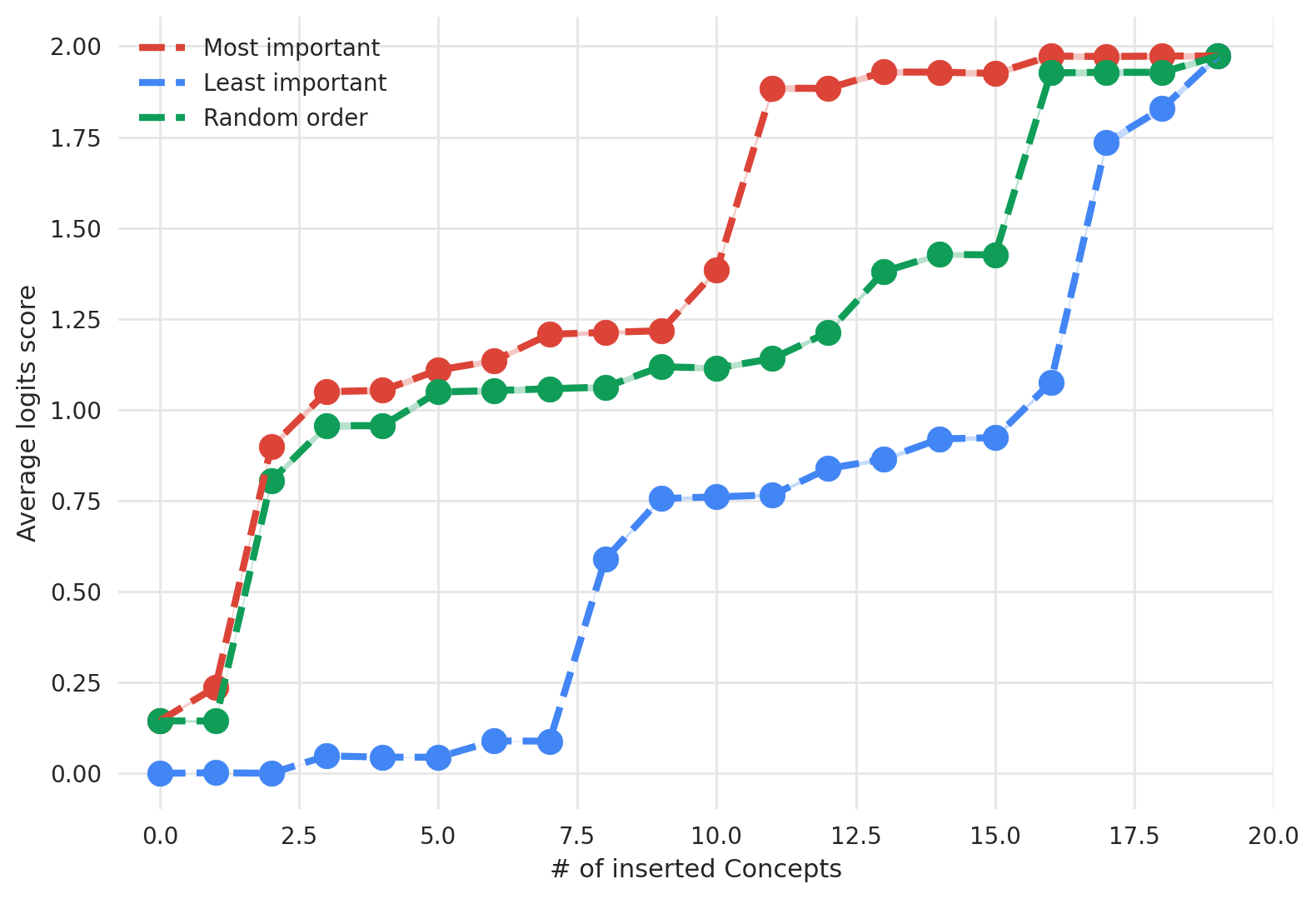}
    \caption{(Upper) Deletion curve for RoBERTa on \textit{IMDB Reviews} (lower is better). (Lower) Insertion
curve for RoBERTa on \textit{IMDB Reviews} (higher is better).}
    \label{fig:fidelity}
\end{figure}

In Fig.~\ref{fig:fidelity}, we showcase the plots for these two fidelity metrics on the \textit{IMDB Reviews} %and on the \textit{Amazon Electronics} 
dataset. %We observe that the concepts are indeed important for the model's predictions in both cases, but that those extracted on for the latter are more faithful to the model than on the first one. We surmise that the task and the specific training may have a considerable influence on the quality of the concepts that the model uses for predicting, and on our method's ability to extract and quantify their importance. Investigating why this happens is left for future works.
We observe that the concepts are indeed important for the model's predictions. In the both plots, the curve corresponding to the concept ranked in order of importance according to our Sobol method is better than a random ranking of these concepts, and much better than if we had taken the order of Sobol importance in reverse. In particular, to obtain statistically significant results, we took 10 sets of 10k reviews, and computed the mean and standard deviation values for both of the metrics.

\subsection{Qualitative evaluation}
\begin{figure}[t]
    \centering
    \includegraphics[width=1\linewidth]
    {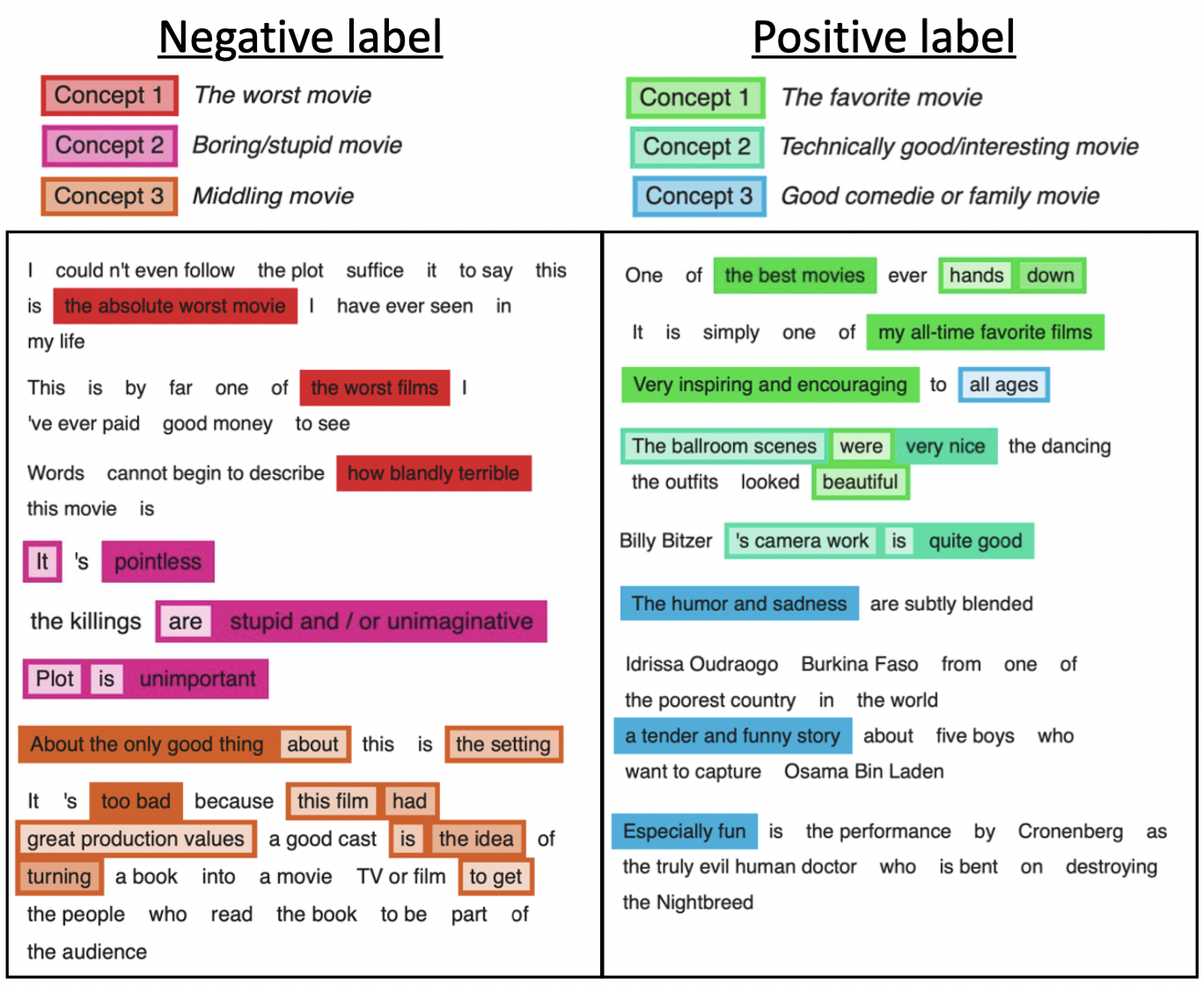}
    \caption{Concepts generated with $l = 20$ for a few sentences taken from IMDB reviews. The colored elements are those important for the concept of the corresponding color (calculated with part \textit{(iii)} of our method). The more colorful the element, the more important it is for the concept (continuously). We have selected the 3 most important concepts for each label (see Fig.~\ref{fig:importanceconcept_roberta}). The name of the concept is chosen manually in view of the important elements corresponding to the concepts.}
    \label{fig:examples}
\end{figure}

%To present qualitative results, we used the IMDB dataset \cite{maas-EtAl:2011:ACL-HLT2011}. The IMDB dataset is a collection of 50K movie reviews from the Internet Movie Database (IMDB) website. For each review, IMDB specifies whether it is positive or negative (the label). The dataset is balanced, with 25K positive and 25K negative reviews. We used a RoBERTa model to predict the label from the reviews.

A model with a good accuracy like RoBERTa gives very good explanations.  Others like LSTM (see appendix \ref{apdx:LSTM}) do not do so well and do not yield good explanations.  This is not a surprise; if the model predicts badly, necessarily the concepts it uses to predict will be bad. Similarly, if the model is very basic, it uses simple concepts to predict. 
The reviews in IMDB are also well written, so it is more comfortable to analyse sentences and words to properly call the concepts found by the NMF. 

In Fig.~\ref{fig:examples}, we can see the 3 most important concepts for each label class. Each of its concepts "\textit{the favorite movie}", "\textit{technically good/interesting movie}", "\textit{good comedie of family movie}" for the positive class or "\textit{the worst movie}", "\textit{middling movie}", "\textit{boring/stupid movie}" for the negative class are ideas that seem natural and which structures our vision of why a film would be positive or negative.

\section{Conclusion}

In this paper, we revisited concept-based explainability techniques and presented COCKATIEL, a post-hoc, model agnostic method capable of generating meaningful and faithful explanations for NLP models trained on classification tasks.
The method has three parts: \textit{(i) a concept part}, using Non-Negative Matrix Factorization to discover the concept, \textit{(ii) a ranking part}, using Total Sobol indices to measure the influence of each concept, and \textit{(iii) an interpretable elements part}, using a black-box attribution method to quantify the impact of each element out of each concept.

We measured COCKATIEL's ability to discover concepts that align with those humans and obtained better scores than state-of-the-art methods. We demonstrated that in addition to generating meaningful concepts for humans, these explanations are faithful to the models. Finally, we gave some qualitative examples of explanations for different models to understand the method "in practice".

\section*{Limitations}

We have demonstrated that COCKATIEL is capable of generating meaningful explanations that align with human concepts, and that they tend to explain rather faithfully the model. 
%However, we experienced a certain variance in the quality of the fidelity curves, namely on the \textit{Beer Reviews} dataset, where we observed important ``bumps'' quite far from the concepts that were ranked as the most important. This could serve as a signal that the there's a certain \agus{pas sur de vouloir mettre ceci...}

%It would also be interesting to study the evolution in the complexity of concepts as we get deeper in the model. We have shown that the concepts extracted from transformer models' classification heads can be meaningful to humans, but by looking at shallower blocks, we might be able to bring to light how these complex models deal with the compositionality of abstract concepts.

The concepts extracted of NMF are abstract and we interpret them using part 3 of the method. However, for the interpretation, we rely on our own understanding of the concept linked to the examples of words or clauses associated with the concept. This part therefore requires human supervision and will not be identical depending on who is looking. One way to add some objectivity to this concept labeling task would be to leverage topic modeling models to find a common theme to each concept.

In addition, $\bm{\tau}_1$ and $\bm{\tau}_2$ were chosen empirically to allow for an adequate concept complexity/human understandability trade-off in our examples. We recognize that this choice might not be optimal in every situation, as more complex concept may be advantageous in some cases, and more easily understandable ones, in others. We surmise that this choice might also depend on the amount of concepts and on the model's expressivity.

Finally, we have studied the meaningfulness and fidelity of our generated concepts, but ideally, the simulatability should also be tested. This property measures the explanation's capacity to help humans predict the model's behavior, and has recently caught the attention of the XAI community~\cite{fel2021cannot,shen2020useful,nguyen2018comparing,hase2020evaluating}. We leave this analysis for future works.

\section*{Ethics Statement}

This work contributes to the field of explainability. This field has strong links with the field of fairness, because explaining a model makes it possible to understand its biases. Transformers are a type of model that are little studied in explainability and yet it is widely used. COCKATIEL is a tool to explain transformers and therefore avoid using biased models against the minority.

It is important to remark that this need for understanding automatic decisions start being enforced by Law, as for instance by the so-called \emph{AI act}\footnote{\url{https://eur-lex.europa.eu/legal-content/EN/TXT/?uri=celex\%3A52021PC0206}} of the European Union. As a consequence, companies need to ensure that they are deploying algorithms which are free of harmful biases and that the explanations that they're obligated to issue are easily understandable by employees and end-users alike.  

\section*{Acknowledgements}
We thank the ANR-3IA Artificial and Natural Intelligence Toulouse Institute (ANITI) funded by the ANR-19-PI3A-0004 grant for research support.
We also thank the reviewers for their insightful comments.
This work was conducted as part of the DEEL\footnote{\url{https://www.deel.ai}} project. 


\bibliography{main}
\bibliographystyle{main}

\appendix

\section{Sobol technique in details}\label{apdx:sobol}
%%reprendre
%We propose to formally derive the Sobol indices for the estimation of the importance of concepts. 
%%

Let $(\Omega, \mathcal{A}, \mathbb{P})$ be a probability space of possible concept perturbations. To build these concept perturbations, we use $\boldsymbol{M}=\left(M_1, \ldots, M_r\right) \in \mathcal{M} \subseteq[0,1]^r$, i.i.d. stochastic masks on the original vector of concept coefficients $\widehat{U} \in \mathbb{R}^r$. We define concept perturbation $U=\boldsymbol{\pi}(\widehat{\boldsymbol{U}}, \boldsymbol{M})$ with the perturbation operator $\pi(\tilde{\boldsymbol{U}}, \boldsymbol{M})=\tilde{\boldsymbol{U}} \odot \boldsymbol{M}+(\mathbf{1}-\boldsymbol{M}) \mu$ with $\odot$ the Hadamard product and $\mu \in \mathbb{R}$ a baseline value, here zero.

%Concretely, to create our concept perturbation we consider the inpainting function as our perturbation operator (as in \cite{fel2021sobol, petsiuk2018rise, ribeiro2016should}): $\pi(\tilde{\boldsymbol{U}}, \boldsymbol{M})=\tilde{\boldsymbol{U}} \odot \boldsymbol{M}+(\mathbf{1}-\boldsymbol{M}) \mu$ with $\odot$ the Hadamard product and $\mu \in \mathbb{R}$ a baseline value, here zero. 

%For the sake of notation, we will note $f: \mathcal{A} \rightarrow \mathbb{R}$ the function mapping a random concept perturbation $U$ from the layer $l$ to the output. 

We denote the set $\mathcal{U}=\{1, \ldots, r\}$, $\boldsymbol{u}$ a subset of $\mathcal{U}$, its complementary $\sim \boldsymbol{u}$ and $\mathbb{E}(\cdot)$ the expectation over the perturbation space. We define $\boldsymbol{c}: \mathcal{A} \rightarrow \mathbb{R}$, the classification function and we assume that $\boldsymbol{c} \in \mathbb{L}^2(\mathcal{A}, \mathbb{P})$ i.e. $|\mathbb{E}(\boldsymbol{c}(\boldsymbol{U}))|<+\infty$.

The Hoeffding decomposition gives $\boldsymbol{c}$ in function of summands of increasing dimension, denoting $\boldsymbol{c}_{\boldsymbol{u}}$ the partial contribution of the concepts $\boldsymbol{U}_{\boldsymbol{u}}=\left(U_i\right)_{i \in \boldsymbol{u}}$ to the score $\boldsymbol{c}(\boldsymbol{U})$ :

\begin{equation}
\label{eq:hoeffding}
\begin{aligned}
\boldsymbol{c}(\boldsymbol{U}) & =\boldsymbol{c}_{\varnothing} \\
& +\sum_i^r \boldsymbol{c}_i\left(U_i\right) \\
& +\sum_{1 \leqslant i<j \leqslant r} \boldsymbol{c}_{i, j}\left(U_i, U_j\right)+\cdots \\
& +\boldsymbol{c}_{1, \ldots, r}\left(U_1, \ldots, U_r\right) \\
& =\sum_{\boldsymbol{u} \subseteq \mathcal{U}} \boldsymbol{c}_{\boldsymbol{u}}\left(\boldsymbol{U}_{\boldsymbol{u}}\right)
\end{aligned}
\end{equation}

Eq.~\ref{eq:hoeffding} consists of $2^r$ terms and is unique under the orthogonality constraint:
$$
\quad \mathbb{E}\left(\boldsymbol{c}_{\boldsymbol{u}}\left(\boldsymbol{U}_{\boldsymbol{u}}\right) \boldsymbol{c}_{\boldsymbol{v}}\left(\boldsymbol{U}_{\boldsymbol{v}}\right)\right)=0,  \forall(\boldsymbol{u}, \boldsymbol{v}) \subseteq \mathcal{U}^2 \text { s.t. } \boldsymbol{u} \neq \boldsymbol{v}
$$
Moreover, thanks to orthogonality, we have $\boldsymbol{c}_{\boldsymbol{u}}\left(\boldsymbol{U}_{\boldsymbol{u}}\right)=\mathbb{E}\left(\boldsymbol{c}(\boldsymbol{U}) \mid \boldsymbol{U}_{\boldsymbol{u}}\right)-\sum_{\boldsymbol{v} \subset \boldsymbol{u}} \boldsymbol{c}_{\boldsymbol{v}}\left(\boldsymbol{U}_{\boldsymbol{v}}\right)$ and we can write model variance as:

\begin{equation}
\label{eq:var}
    \begin{aligned}
\mathbb{V}(\boldsymbol{c}(\boldsymbol{U})) & =\sum_i^r \mathbb{V}\left(\boldsymbol{c}_i\left(U_i\right)\right) \\
& +\sum_{1 \leqslant i<j \leqslant r} \mathbb{V}\left(\boldsymbol{c}_{i, j}\left(U_i, U_j\right)\right) \\
& +\ldots+\mathbb{V}\left(\boldsymbol{c}_{1, \ldots, r}\left(U_1, \ldots, U_r\right)\right) \\
& =\sum_{\boldsymbol{u} \subseteq \mathcal{U}} \mathbb{V}\left(\boldsymbol{c}_{\boldsymbol{u}}\left(\boldsymbol{U}_{\boldsymbol{u}}\right)\right)
\end{aligned}
\end{equation}

Eq.~\ref{eq:var} allows us to write the influence of any subset of concepts $\boldsymbol{u}$ as its own variance. This yields, after normalization by $\mathbb{V}(\boldsymbol{c}(\boldsymbol{U}))$, the general definition of Sobol' indices.

\begin{definition}\textit{Sobol indices \cite{sobol1993sensitivity}.}
The sensitivity index $\mathcal{S}_{\boldsymbol{u}}$ which measures the contribution of the concept set $\boldsymbol{U}_{\boldsymbol{u}}$ to the model response $f(\boldsymbol{U})$ in terms of fluctuation is given by:
\begin{equation}
\label{eq:sobol}
    \begin{aligned}
\mathcal{S}_{\boldsymbol{u}} & =\frac{\mathbb{V}\left(\boldsymbol{c}_{\boldsymbol{u}}\left(\boldsymbol{U}_{\boldsymbol{u}}\right)\right)}{\mathbb{V}(\boldsymbol{c}(\boldsymbol{U}))} \\
& =\frac{\mathbb{V}\left(\mathbb{E}\left(\boldsymbol{c}(\boldsymbol{U}) \mid \boldsymbol{U}_{\boldsymbol{u}}\right)\right)-\sum_{\boldsymbol{v} \subset \boldsymbol{u}} \mathbb{V}\left(\mathbb{E}\left(\boldsymbol{c}(\boldsymbol{U}) \mid \boldsymbol{U}_{\boldsymbol{v}}\right)\right)}{\mathbb{V}(\boldsymbol{c}(\boldsymbol{U}))}
\end{aligned}
\end{equation}
\end{definition}

Sobol indices provide a numerical assessment of the importance of various subsets of concepts in relation to the model's decision-making process. Thus, we have: $\sum_{\boldsymbol{u} \subseteq \mathcal{U}} \mathcal{S}_{\boldsymbol{u}}=1$.

%Furthermore, the framework of Sobol' indices enables us to easily capture higher-order interactions between features. 
Additionally, the use of Sobol' indices allows for the efficient identification of higher-order interactions between features. Thus, we can view the Total Sobol indices defined in \ref{eq:total_sobol} as the sum of of all the Sobol indices containing the concept $i: \mathcal{S}_{T_i}=\sum_{\boldsymbol{u} \subseteq \mathcal{U}, i \in \boldsymbol{u}} \mathcal{S}_{\boldsymbol{u}}$. 

\section{Implementation Details}\label{apdx:finetuning}

We trained 3 different models. For each model, we performed a single run and we split datasets in 70\% for train, 10\% for validation and 20\% for test. 

\subsection{Trained RoBERTa on Beer dataset}
We used a RoBERTa base pretrained on hugging face by \citet{DBLP:journals/corr/abs-1907-11692} (all the information on the pretrain can be found in the paper). The model was pretrained on the reunion of five datasets:
\begin{itemize}
    \item BookCorpus \cite{moviebook}, a dataset containing 11,038 unpublished books;
    \item English Wikipedia (excluding lists, tables and headers) ; 
    \item CC-News \cite{10.1145/3340531.3412762}, a dataset containing 63 millions English news articles crawled between September 2016 and February 2019 ;
    \item OpenWebText \cite{radford2019language}, an opensource recreation of the WebText dataset used to train GPT-2 ;
    \item Stories \cite{trinh2018simple} a dataset containing a subset of CommonCrawl data filtered to match the story-like style of Winograd schemas.
\end{itemize}

We then trained the model on Beer dataset. The model was trained on 2 GPUs for 10 epochs with a batch size of 32 and a sequence length of 512. The optimizer was AdamW with a learning rate of 1e-5, $\beta_1 = 0.9$, $\beta_2 = 0.98$, and $\epsilon = 1e6$ %, learning rate warmup for 0 steps and linear decay of the learning rate after.

\subsection{Trained RoBERTa on IMDB dataset}

We used a RoBERTa model already fine-tuned on IMDB from hugging face.
This model used the pre-training presented above, we fine-tuned it with 2 epochs, a batch size of 16, and an Adam optimizer with a learning rate of 2e-5, $\beta_1 = 0.9$, $\beta_2 = 0.999$ and $\epsilon = 1e-8$.

\subsection{Trained LSTM on IMDB dataset}

We created our LSTM with:
\begin{lstlisting}
SentimentRNN(
  (embedding): Embedding(1001, 512)
  (lstm): LSTM(512, 128, 
  num_layers=4, batch_first=True, 
  bidirectional=True)
  (dropout): Dropout(p=0.3, 
  inplace=False)
  (fc_1): Linear(in_features=128,
  out_features=128, bias=True)
  (relu): ReLU()
  (fc_2): Linear(in_features=128,
  out_features=2, bias=True)
  (sig): Softmax(dim=1)
)
\end{lstlisting}

Then, we trained it on the IMDB dataset. The model was trained on 2 GPUs for 5 epochs with a batch size of 128 and a sequence length of 512. The optimizer was Adam with a learning rate of 1e-4.

\section{LSTM example}\label{apdx:LSTM}

LSTMs are much less complex than RoBERTa, and as such, we can expect them to leverage less and much simpler concepts for their predictions.
% LSTM is a less complex model than RoBERTa and does not use as many concepts.

In particular, COCKATIEL identified 3 concepts that monopolized the importance score for each class on the RoBERTa model. For the positive class, we had "\textit{the favorite movie}", "\textit{technically good/interesting movie}" and "\textit{good comedie or family movie}".
For the negative class, we also had "\textit{the worst movie}", "\textit{middling movie}" and "\textit{boring movie}".
% With RoBERTa, COCKATIEL detected 3 fundamental concepts for prediction for each class. For the positive class, we had "\textit{the favorite movie}", "\textit{technically good/interesting movie}" and "\textit{good comedie of family movie}".
% For the negative class, we also had "\textit{the worst movie}", "\textit{middling movie}" and "\textit{boring movie}".

In contrast, in the case of the LSTM (see figure \ref{fig:visu_annexeLSTM}), COCKATIEL detected a single important concept per predicted class. For the positive class, this concept encompasses  \textit{the positive language elements} mostly, and for the negative class, \textit{the negative elements}. This is a much more basic view of the review classification problem, and COCKATIEL allows us to confirm our intuitions about the richness of the embedding learned by the LSTM.
% Our explanations given by COCKATIEL confirm this.

\begin{figure}[t]
    \centering
\includegraphics[width=1\linewidth]
{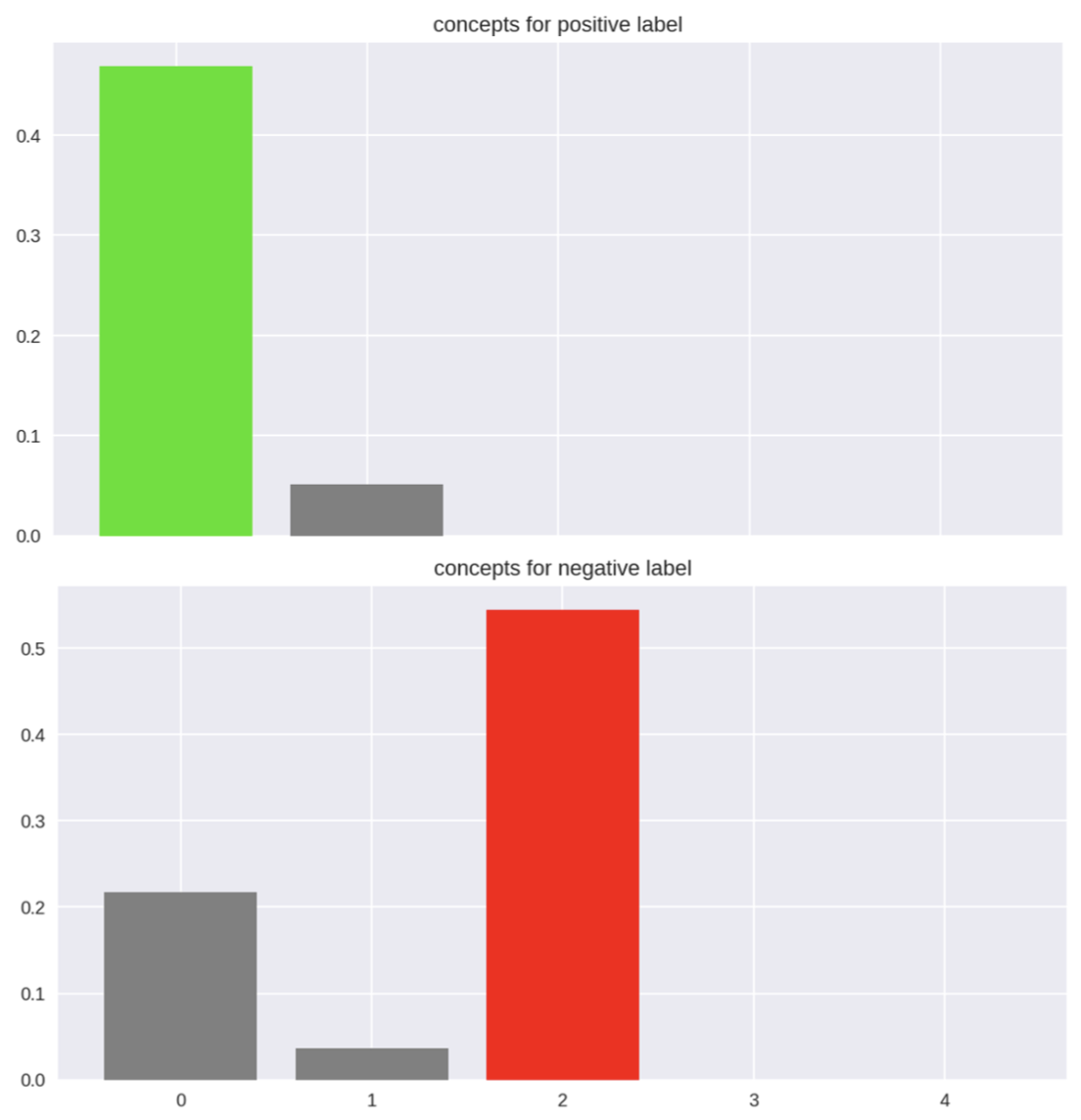}
    \caption{\textbf{Concept importance: } The global influence of the NMF concepts on the predictions on LSTM Model is then measured using Sobol indices. We have different concepts for each class (positive and negative label).}
\label{fig:importanceconcept_LSTM}
\end{figure}

\begin{figure}[t]
    \centering
\includegraphics[width=1\linewidth]
{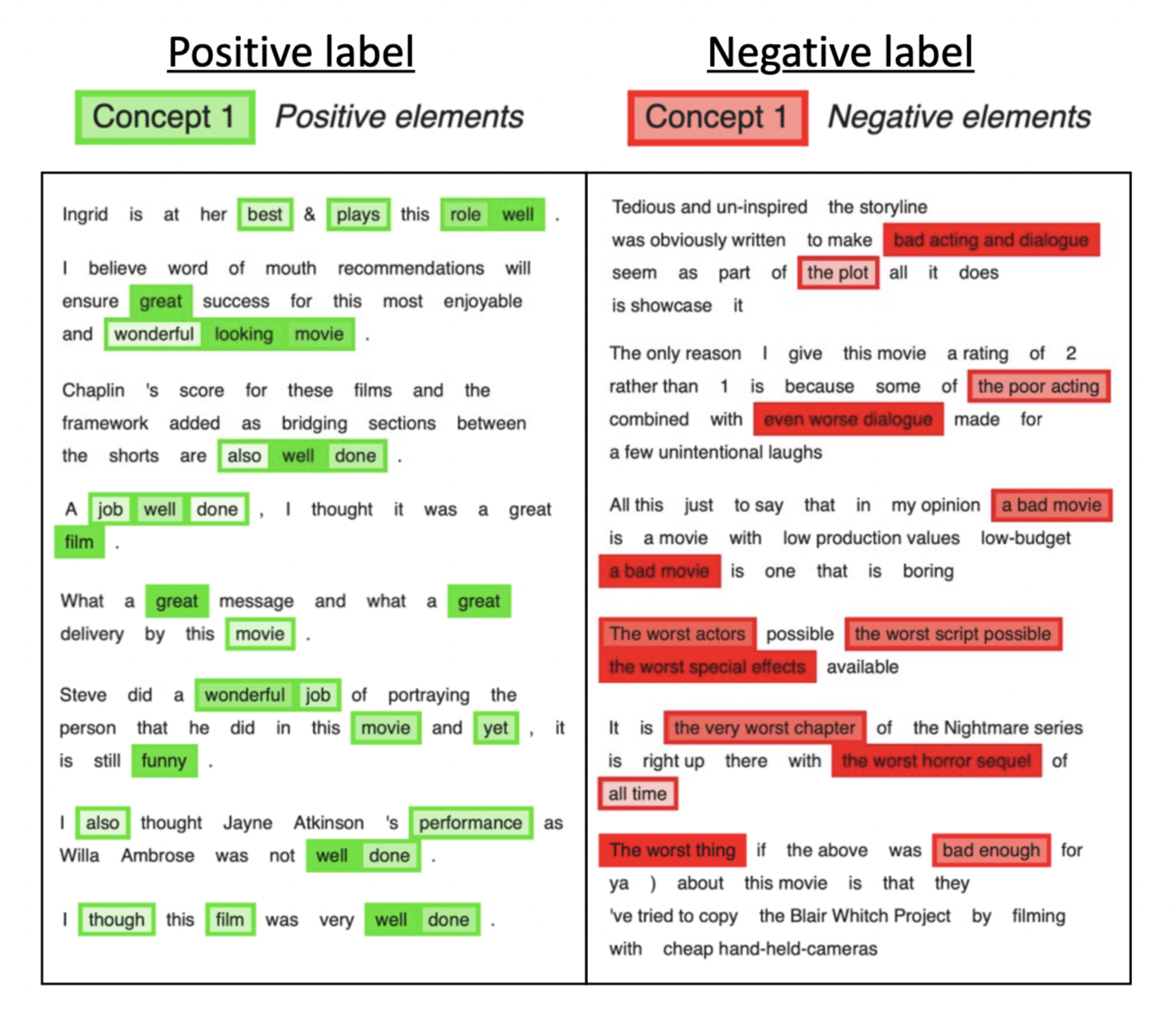}
    \caption{Concepts generated for a LSTM model with $l = 5$ for a few sentences out of  IMDB reviews. The colored elements are those important for the concept of the corresponding color (calculated with part \textit{(iii)} of our method). The more colorful the element, the more important it is for the concept (continuously). We have selected the most important concept for each label (see Fig.~\ref{fig:importanceconcept_LSTM}). The name of the concept is chosen manually in view of the important elements corresponding to the concepts.}
\label{fig:visu_annexeLSTM}
\end{figure}

\section{Other examples of COCKATIEL explanations for RoBERTa}\label{apdx:examples_sup}

\begin{figure}[t]
    \centering
    \includegraphics[width=1\linewidth]
    {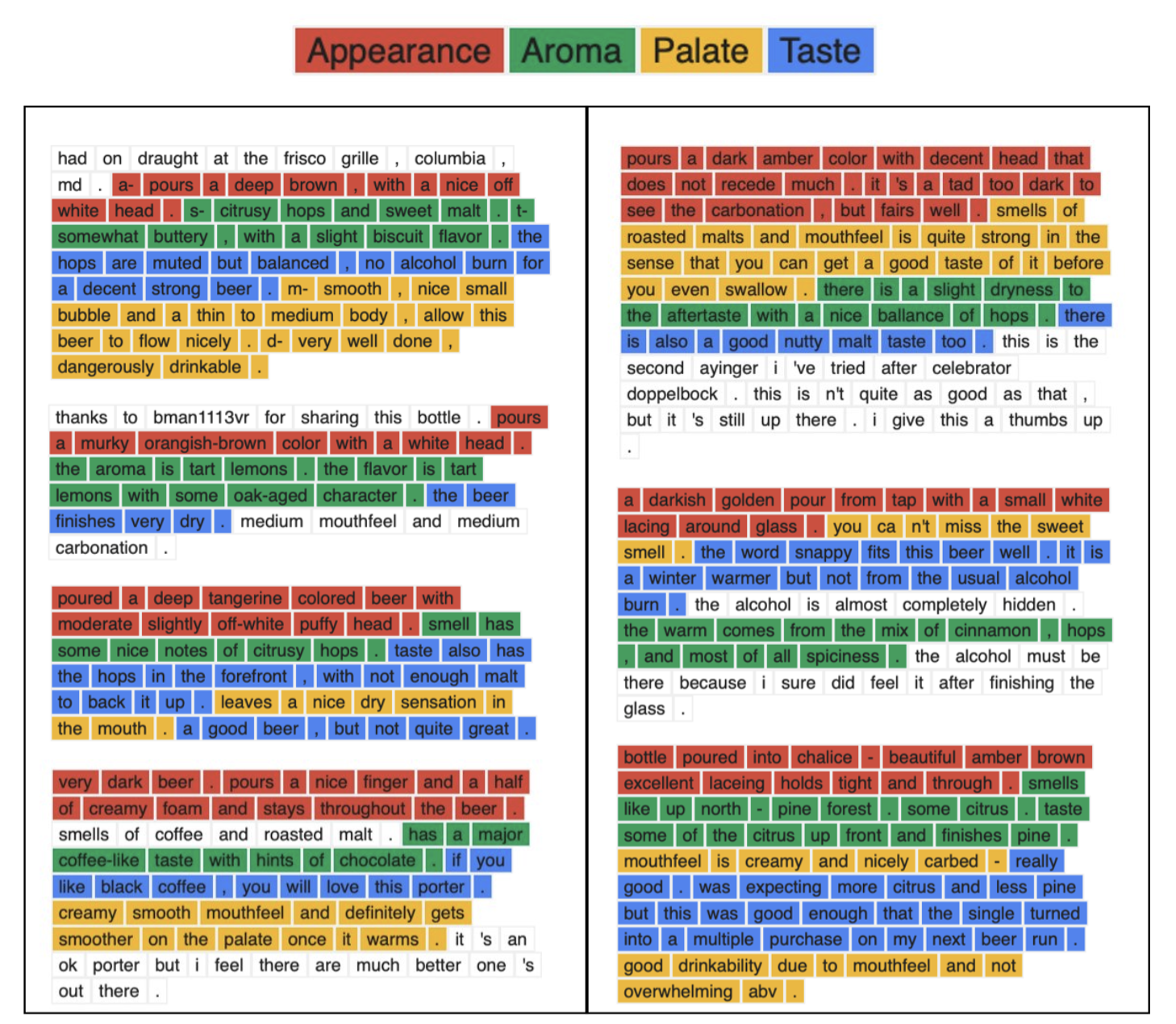}
    \caption{Concepts generated with $l = 20$ for some beer reviews with RoBERTa model. The color depicts the aspects for each annotate concept. COCKATIEL is trained only on the label and we use the NMF part of the method to find annotate concepts.}
    \label{fig:beerexamplesannexe}
\end{figure}

\begin{figure}[t]
    \centering
\includegraphics[width=1\linewidth]
{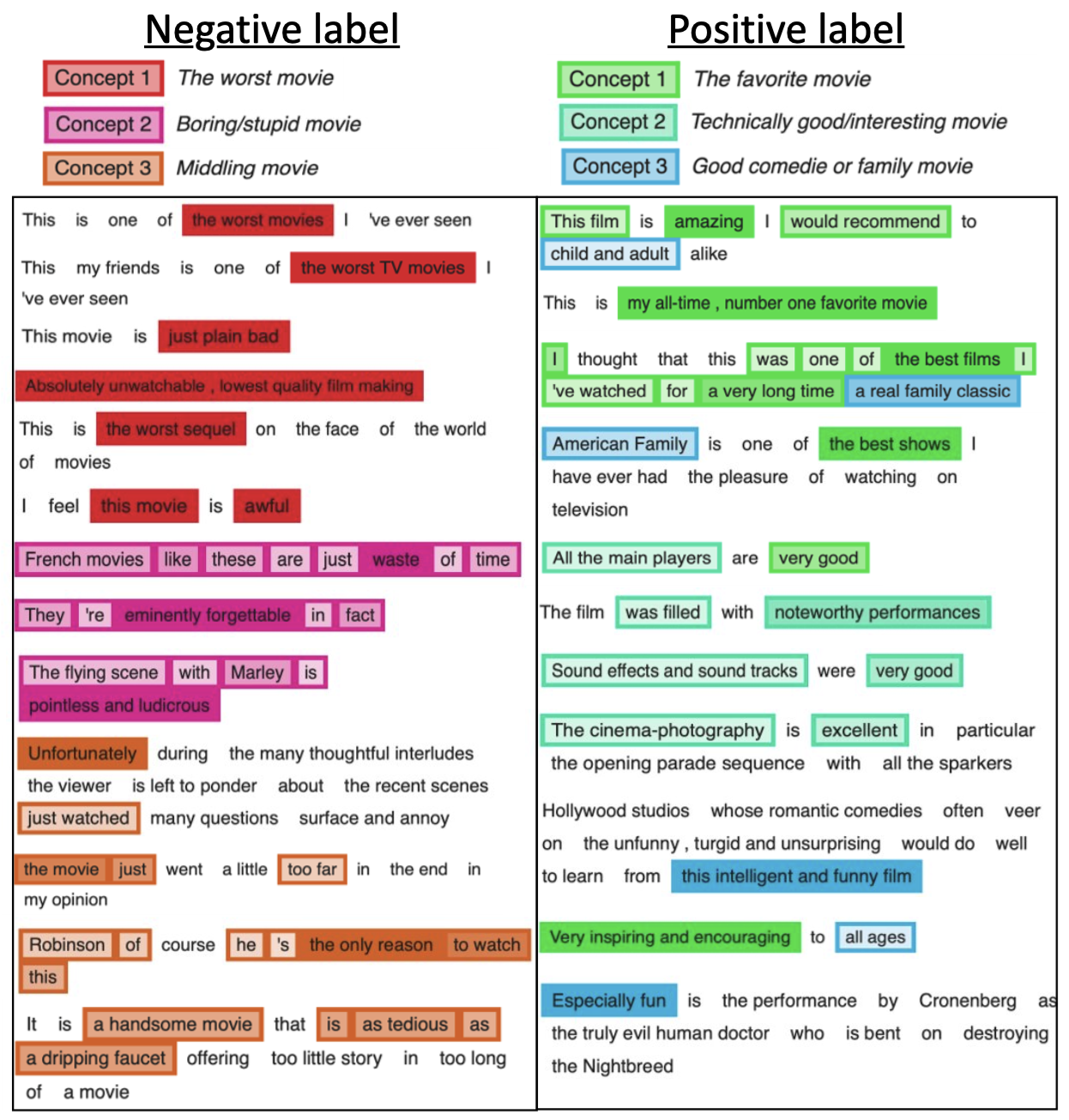}
    \caption{Concepts generated for a RoBERTa model with $l = 20$ for a few sentences taken out of IMDB reviews. The colored elements are those important for the concept of the corresponding color (calculated with part \textit{(iii)} of our method). The more colorful the element, the more important it is for the concept (continuously). We have selected the 3 most important concepts for each label (see Fig.~\ref{fig:importanceconcept_roberta}). The name of the concept is chosen manually in view of the important elements corresponding to the concepts.}
    \label{fig:examplesannexe}
\end{figure}

\begin{figure}[t]
    \centering
\includegraphics[width=1\linewidth]
{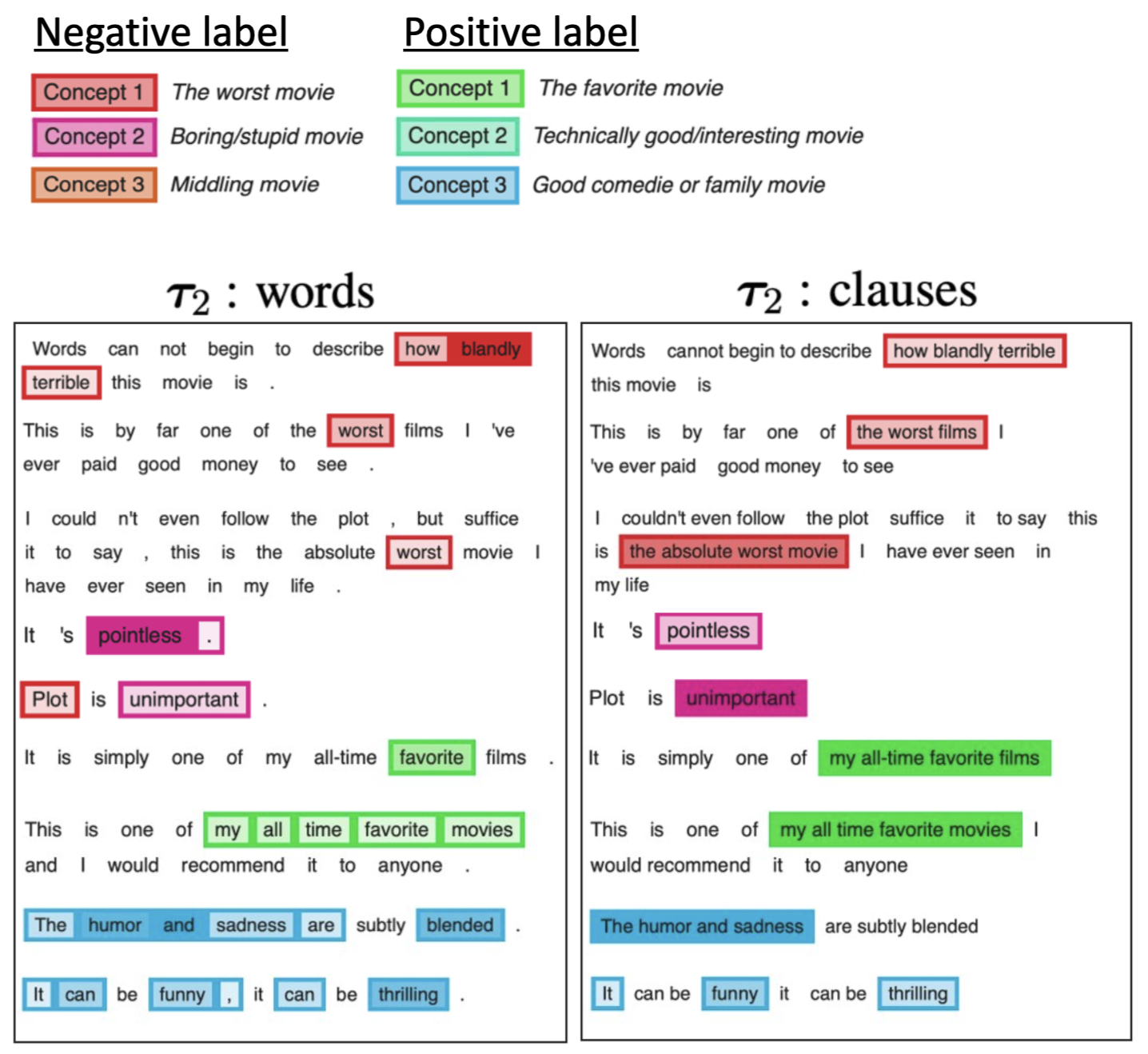}
    \caption{Concepts generated for a RoBERTa model with $l = 20$ for a few sentences taken out of IMDB reviews. 
    The excerpts chosen by an excerpt-extraction function $\bm{\tau}_1$ are sentences for both (so, we have same concepts). The colored elements are those that are considered to be the most important for the concept of the corresponding color (calculated with part \textit{(iii)} of our method). We compare the visualisations of the same sentences with two different excerpt-extraction functions $\bm{\tau}_2$:   words (on the left) and clauses (on the right). 
    We split the text into clauses for occlusion using the fair library’s SequenceTagger implementation.}
    \label{fig:visu_tau}
\end{figure}

\end{document}